\documentclass[journal]{IEEEtai}

\usepackage[colorlinks,urlcolor=blue,linkcolor=blue,citecolor=blue]{hyperref}

\usepackage{color,array}

\usepackage{graphicx}
\usepackage{amssymb}
\usepackage{multirow}
\usepackage{multicol}
\usepackage{makecell}
\usepackage[table,xcdraw]{xcolor}
\usepackage{cite}
\usepackage{graphicx}
\usepackage{float} 
\usepackage{subfigure}

\usepackage{CJKutf8}

\begin{document}
\begin{CJK}{UTF8}{gbsn}
\title{A Survey on Robotic Manipulation of Deformable Objects: Recent Advances, Open Challenges and New Frontiers} 


\author{Feida Gu, Yanmin Zhou, Zhipeng Wang, Shuo Jiang, Bin He
\thanks{Research supported by the National Natural Science Foundation of China under Grant 61825303, Grant 51975415, Grant 62088101; in part by the Shanghai Municipal Science and Technology Major Project under Grant 2021SHZDZX0100; in part by the Shanghai Municipal Commission of Science and Technology Project under Grant 22ZR1467100.(Corresponding author: Zhipeng Wang.)}
\thanks{Feida Gu is with Shanghai Research Institute for Intelligent Autonomous Systems, Tongji University, Shanghai 201804, China, with National Key Laboratory of Autonomous Intelligent Unmanned Systems, Shanghai 201210, China, and also with the Frontiers Science Center for Intelligent Autonomous Systems, Shanghai 200120, China (e-mail: feidagu@tongji.edu.cn).}
\thanks{Yanmin Zhou, Zhipeng Wang, Shuo Jiang, Bin He are with the Department of Control Science and Engineering, Tongji University, Shanghai 201804, China, with National Key Laboratory of Autonomous Intelligent Unmanned Systems, Shanghai 201210, China, and also with the Frontiers Science Center for Intelligent Autonomous Systems, Shanghai 200120, China (e-mail: yanmin.zhou@tongji.edu.cn; wangzhipeng@tongji.edu.cn; jiangshuo@tongji.edu.cn; hebin@tongji.edu.cn).}
}
\maketitle

\begin{abstract}
Deformable object manipulation (DOM) for robots has a wide range of applications in various fields such as industrial, service and health care sectors. However, compared to manipulation of rigid objects, DOM poses significant challenges for robotic perception, modeling and manipulation, due to the infinite dimensionality of the state space of deformable objects (DOs) and the complexity of their dynamics. The development of computer graphics and machine learning has enabled novel techniques for DOM. These techniques, based on data-driven paradigms, can address some of the challenges that analytical approaches of DOM face. However, some existing reviews do not include all aspects of DOM, and some previous reviews do not summarize data-driven approaches adequately. In this article, we survey more than 150 relevant studies (data-driven approaches mainly) and summarize recent advances, open challenges, and new frontiers for aspects of perception, modeling and manipulation for DOs. Particularly, we summarize initial progress made by Large Language Models (LLMs) in robotic manipulation, and indicates some valuable directions for further research. We believe that integrating data-driven approaches and analytical approaches can provide viable solutions to open challenges of DOM.
\end{abstract}

\begin{IEEEImpStatement}
The manipulation of deformable objects is very common in industry or daily life. The applications of deformable object manipulation in industry can create enormous economic benefits, and the applications in daily life can improve quality of life, especially for people with disabilities and the elderly. Robotic manipulation tasks such as food handling, assistive dressing or clothing folding require open problems of perception, modeling and manipulation of deformable objects to be solved. Recent advances in machine learning have addressed some of the limitations of traditional approaches. Combining analytical and data-driven approaches holds the potential to bring the manipulation capabilities of robots close to that of humans. This paper systematically reviews various approaches to robotic manipulation of deformable objects, providing a quick reference for researchers and practitioners in different fields. Additionally, the paper provides insights into open research areas in the field and points out promising future research such as LLMs for robotic manipulation.
\end{IEEEImpStatement}


\section{Introduction}

\IEEEPARstart {T}{he} ability to manipulate objects is an essential feature of robots, as the definition of a robot requires it to exert influence on its surroundings. Over the past decades, research on robot manipulation has progressed considerably, aiming to utilize cost-effective robotic arms and end-effectors to directly interact with the world and accomplish their objectives \cite{kroemer2021review}.

Object rigidity is a common assumption in robot manipulation \cite{zhu2022challenges}. However, this assumption does not hold for all objects and scenarios, as any object can deform under the application of external forces. The rigidity assumption is only valid when the object deformation is negligible for the manipulation task. However, many objects present non-negligible deformations during robotic manipulation tasks.

\begin{figure*}
\centerline{\includegraphics[width=28pc]{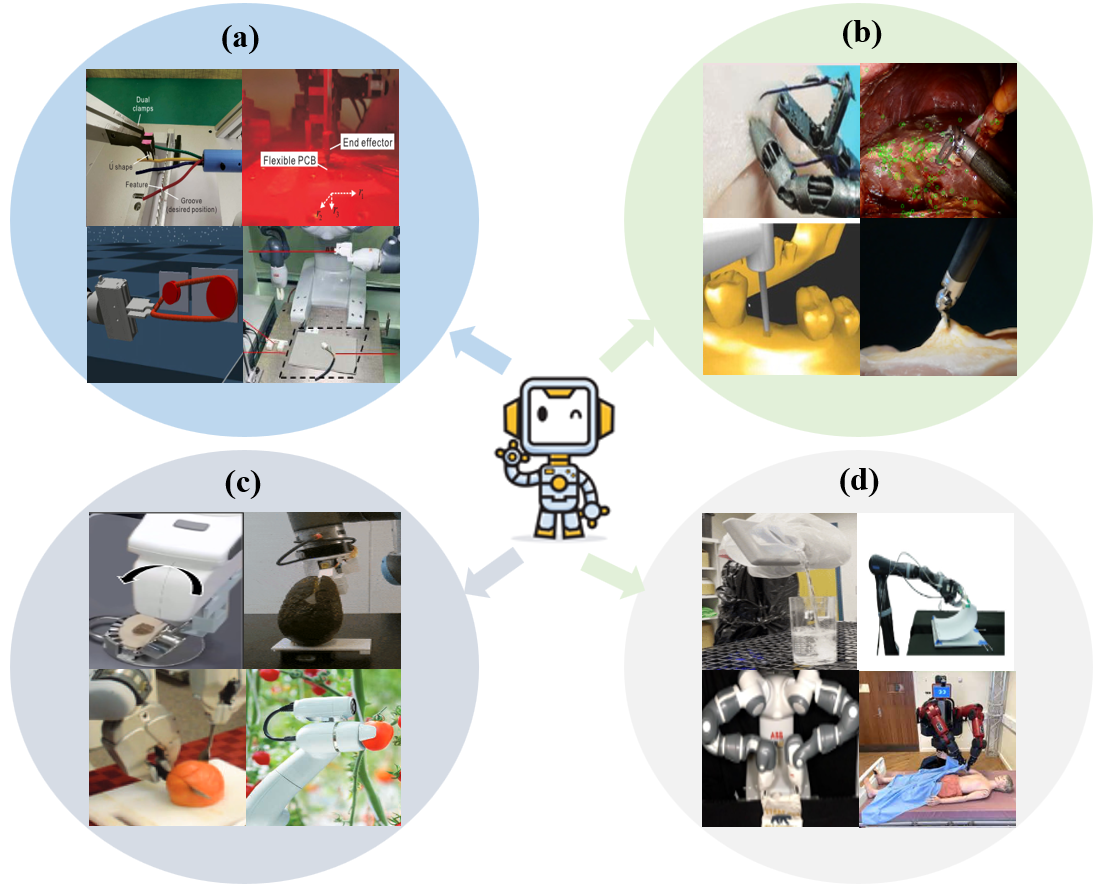}}
\caption{Applications involving DOM. (a) Manufacturing industry \cite{gao2022hierarchical, li2018vision, jin2021trajectory, zhou2020practical} (b) Medical surgery \cite{cao2020sewing, haouchine2013image, chen2018review, liu2021real} (c) Food processing \cite{shi2023robocook,xu2023roboninja,gemici2014learning,li2019factors} (d) Daily living activities \cite{narasimhan2022self, zheng2022autonomous, avigal2022speedfolding, zhang2022learning}}  
\label{applications}
\end{figure*}

Enabling robots to manipulate DOs opens up a wide range of potential applications in various fields such as industrial manufacturing, medical surgery, food processing and elderly care. These applications have the potential to generate substantial economic benefits. Fig. \ref{applications}. gives some application examples in various sectors. In manufacturing plants, robots can manipulate DOs to alleviate physical burden on workers \cite{gao2022hierarchical, li2018vision, jin2021trajectory, zhou2020practical}. In the medical field, robots can assist in various surgeries \cite{cao2020sewing, haouchine2013image, chen2018review, liu2021real}. The use of robots in the food processing can reduce labor costs \cite{shi2023robocook, xu2023roboninja, gemici2014learning, li2019factors}. Robots can assist the elderly and the disabled in their daily living activities \cite{narasimhan2022self, zheng2022autonomous, avigal2022speedfolding, zhang2022learning}.

\begin{figure*}
\centerline{\includegraphics[width=28pc]{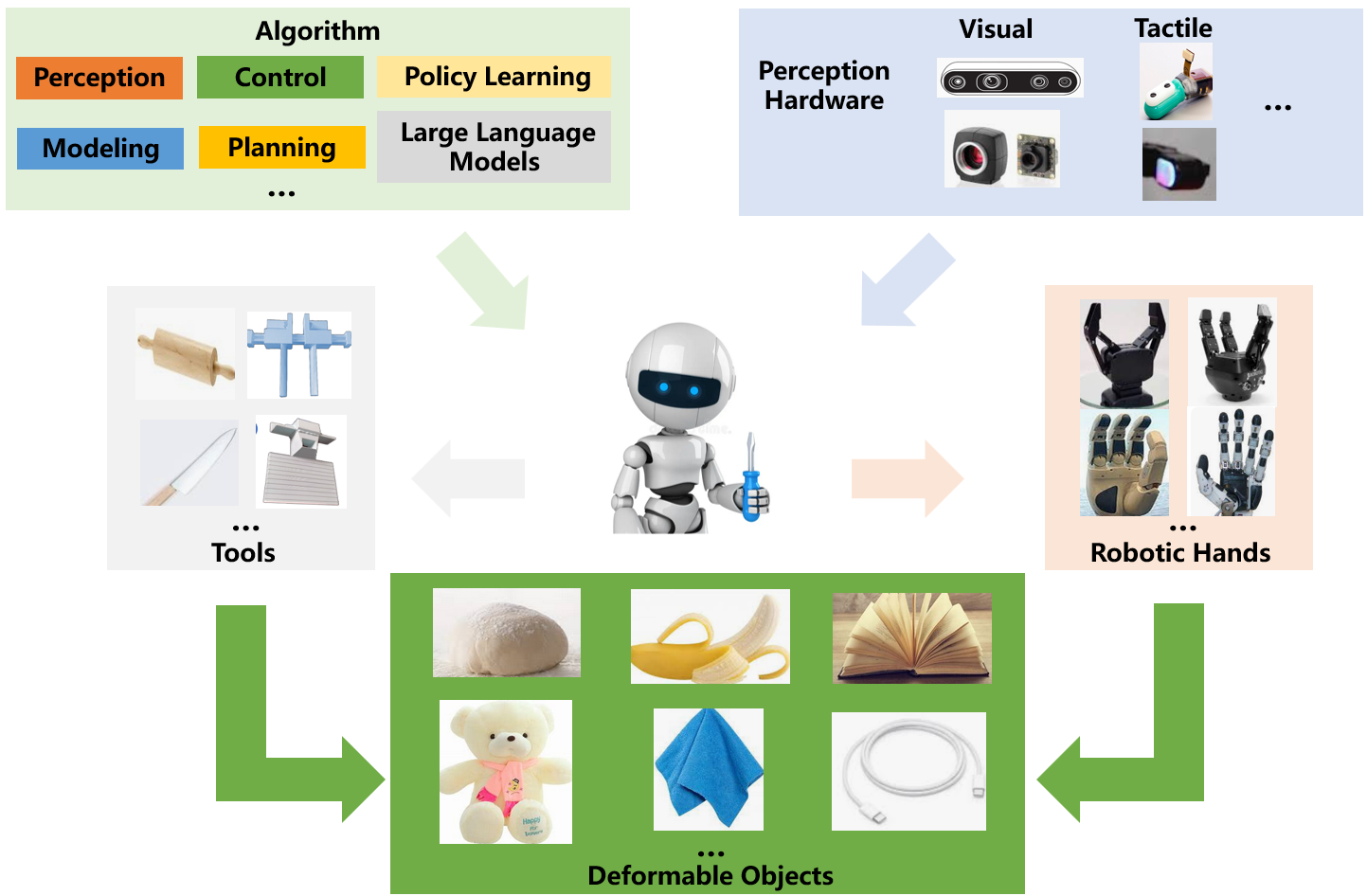}}
\caption{A typical robotic system for handling DOs including robotic hardware, perception hardware, robotic hands, tools used for DOM, algorithms for various functions, etc.}
\label{framework}
\end{figure*}

Despite its importance, DOM has historically been less studied than manipulation of rigid objects since its complexity of perception, modeling and manipulation. When manipulating DOs, the planning strategy of a robotic system (shown in Fig. \ref{framework}.) should not only focus on the motion of a robot, but also need to consider the DOs being manipulated. However, DOs have infinite dimensionality and complex dynamics, making DOM extremely challenging. Recent developments in computer graphics and machine learning provide useful techniques for DOM. Data-driven methods can address some of the limitations of traditional DOM approaches \cite{billard2019trends, yin2021modeling}. 

The review surveyed by Kadi and Terzic \cite{kadi2023data} only focuses on manipulation of clothing using data-driven approaches. Some past reviews \cite{lv2020review, hou2019review, arriola2020modeling} focus on modeling DOs without considering perception and manipulation. The previous reviews \cite{sanchez2018robotic, zhu2022challenges} are comprehensive, but they mainly focus on analytical approaches. The review surveyed by Yin \textit{et al.} \cite{yin2021modeling} concludes a comprehensive review which includes both analytical and data-driven approaches. Our survey mainly focuses on data-driven approaches for perception, modeling and manipulation for DOs, although analytical approaches are also concisely reviewed. In particular, progress in LLMs are discussed in our review. Below, our contribution is concluded specifically. In the perception section, we summarize the perception module in terms of visual, tactile and multimodal, and regard multimodal manipulation datasets and tactile simulators as two key elements for perception of DOM in the future. In the modeling section, we focus on modeling DOs using graph neural networks (GNNs), which is not thoroughly discussed in previous reviews. In the manipulation section, more cutting-edge reinforcement learning (RL) and imitation learning (IL) methods are reviewed. We also discuss current open challenges and point out future research directions for perception, modeling and manipulation respectively. Particularly, we surveyed research about robotic manipulation using LLMs including task definition for manipulation, planning, reward function design and uncertainty alignment, which is not mentioned in previous reviews.
 
The structure of this article is as follows. Section II provides an overview of techniques for sensing DOs, including visual perception and tactile perception. In section III, analytical modeling approaches and data-driven modeling approaches are included. Section IV reviews manipulation methods, including traditional planning and control methods as well as learning-based methods. In Section V, we discuss how LLMs can impact DOM. Finally, Section VI concludes the article.

\section{Perception}
Robots require fast, accurate and multi-modal perception abilities to sense their surroundings, which is a prerequisite for accomplishing complex manipulation tasks \cite{lee2020making}. The aim of perception is to estimate the state of a object $X^\ast$ by solving an optimization problem (Eq. 1) given the observation \textit{O} and the object representation $R(\cdot)$,
\begin{equation}
X^\ast = \mathop{\arg \min} \limits_{X} \||O-R(X)||
\end{equation}
where $X$ is objects’ states.

However, finding accurate and efficient representation of a deformable object (DO) is an open challenge, and solutions are application-specific. Representing DOs as particles is one of commonly used approaches \cite{battaglia2016interaction}. In the perception section of this review, examples use particles to represent DOs, unless otherwise specified.

Since DOs have an infinite-dimensional state space, the perception of deformations becomes highly challenging. Moreover, occlusion and noise are common in unstructured environments, which requires high robustness of state estimation for DOs. 

Visual perception and tactile perception are reviewed in this section. Table. \ref{perceptiontable}. summarizes the primary advantages, limitations and representative literature.

\subsection{Visual perception}
\begin{figure*}
\centerline{\includegraphics[width=28pc]{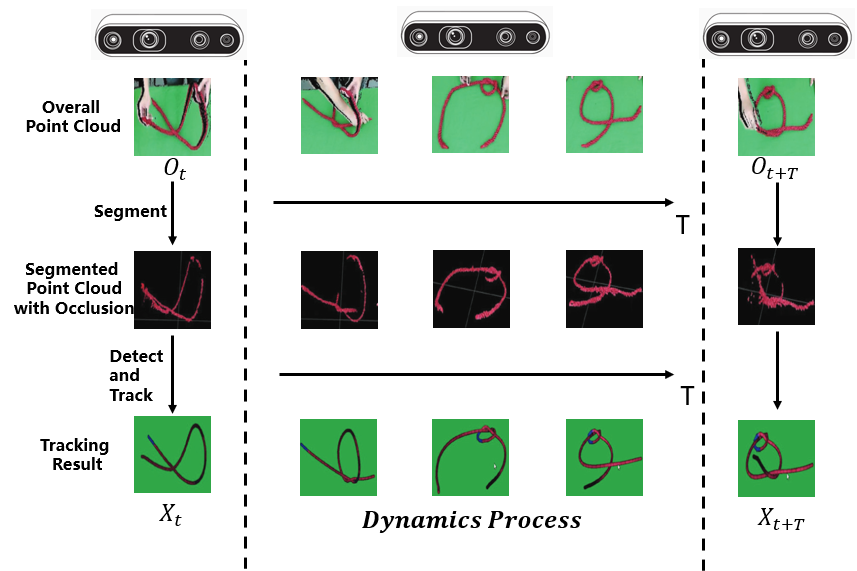}}
\caption{Real-time tracking of DOs (Deformable linear objects (DLOs) as an example \cite{tang2022track}). The first row shows the overall point cloud. The second row shows the segmented point cloud. The third row shows the real-time tracking results.}
\label{visual}
\end{figure*}

A complete vision-based pipeline for estimating the state of a DO consists of three steps including segmentation, detection and tracking. Segmentation aims to separate the DO from the background in an image. Detection refers to the process of estimating the DO’s state from a single image. Tracking is tracking the DO’s state across multiple frames. Fig. \ref{visual}. shows a real-time tracking of DOs.

\subsubsection{Segmentation}
Segmenting DO regions in an image is the first step in the whole processing stream. It is aimed at isolating the DOs from their surroundings. A major challenge of this task is to segment DO instances reliably, accurately and quickly without prior knowledge of the background features and the object count in a scene.

The identification of DOs in simple settings is a solved problem \cite{keipour2022deformable, yan2020self}. In case of complex backgrounds, some advanced DO-specific approaches based on datasets have been proposed \cite{de2018let, ma2023hierarchical, caporali2022ariadne+}. A major obstacle for applying data-driven methods to DO segmentation is the lack of large-scale and high-quality datasets that are publicly available and annotated with labels. To tackle the problem, several methods \cite{denninger2020blenderproc, qiu2016unrealcv} that concentrate on synthetic data generation processes have been proposed. With data generation processes, an advanced method called FASTDLO has been introduced for segmenting DLOs in an image \cite{caporali2022fastdlo}. This technique employs a deep convolutional neural network to perform background segmentation, which produces a binary mask that separates DLOs from the background pixels in an image. FASTDLO has achieved processing speeds of more than 20 FPS in case of complex backgrounds. In addition to DLO segmentation, segmenting fabrics is also studied. Qian \textit{et al.} \cite{qian2020cloth} proposed a neural network that can segment the fabric regions that are suitable for robotic grasping based on the depth image of the scene. The network assigns a semantic label to each pixel, indicating whether it belongs to an outer edge, an inner edge, a corner or none of these categories of the fabric. These labels are then used to estimate the grasping point and direction based on the uncertainty measurement for each outer edge point. The proposed method demonstrated superior performance over the baseline methods on a real robot system that can grasp the edges and corners of cloth in various crumpled configurations. Other studies \cite{dinkel2022wire, zanella2021auto} also focus on obtaining high-quality pixel-level masks of DOs using data-driven methods.

\subsubsection{Detection}
Detection is the process of estimating the positions of the particles which represent DOs in a single frame, using preprocessed sensory data as input.

Estimating the state of DOs from a single-frame point cloud with occlusion is a challenging task. Existing methods, such as the one proposed by Wnun \textit{et al.} \cite{wnuk2020kinematic}, can extract a skeleton curve and 3-D joint positions from the point cloud, but they suffer from occlusion and DOs' diversity. To address these challenges, Lv \textit{et al.} \cite{lv2023learning} proposed a novel dual-branch network architecture that can effectively capture both global and local features of input point cloud for DO estimation. Specifically, the network comprises a PointNet++ encoder, a regression branch for end-to-end estimation and a voting branch for point-to-point refinement. The regression branch concentrates on the global shape of DOs, while the voting branch attends to the local details of DOs. The network also incorporates a fusion module that integrates the outputs of the two branches to produce a smooth and precise estimation of DOs' state.

\subsubsection{Tracking}
Tracking is the process of estimating the deformation of a DO across a sequence of frames. The objective is to generate the state that is consistent with the geometry of a DO at each frame .

Accurate tracking is challenging when objects are partially occluded. Recent advances in computer vision have demonstrated effective methods for DO reconstruction based on various techniques \cite{newcombe2015dynamicfusion, dou2017motion2fusion}. With real-time reconstruction, these methods can dynamically change the tracking model to avoid the occlusion problem. However, they do not meet the criterion of model consistency that is essential for visual-servoing algorithms applied to DOM. DO tracking is also an active research area in surgical robotics. Some studies \cite{collins2016robust, haouchine2013image} have achieved remarkable results in tracking soft tissues in surgical scenarios, but these methods are domain-specific and may not be applicable to other types of objects such as ropes or fabrics. In the domain of robotics, physics simulation is a common technique for tracking DOs that are partially occluded by other objects. This approach has been adopted by several studies \cite{schulman2013tracking, petit2017tracking}. More recent approaches \cite{tang2017state, tang2018framework} have also employed Gaussian Mixture Modeling (GMM) and Coherent Point Drift (CPD \cite{myronenko2010point}) algorithms to generate inputs for their physics simulation engine. Nevertheless, the performance of these algorithms relies on the availability of the physical model of DOs and environmental geometry which are difficult to obtain in unstructured scenarios. Using CPD, Chi and Berenson \cite{chi2019occlusion} developed a framework for tracking DOs in RGBD sequences that can handle occlusion without relying on the physics model of DOs and geometry of the environment. This framework demonstrates higher accuracy than physics-based CPD methods even under occlusion, while maintaining a reasonable run-time.

\subsection{Tactile perception}
The advancement of tactile sensing has stimulated a growing interest among researchers in exploring the applications of tactile technology for DOM. Vision is a sensory modality that mainly obtains global features such as shape and color, whereas tactile sensing is a sensory modality that mainly obtains local features such as friction and texture. Tactile sensing is a valuable modality for manipulation tasks in occluded environments, as it can provide local information about physical properties and interactions of objects involved. 

\begin{figure*} 
\centerline{\includegraphics[width=28pc]{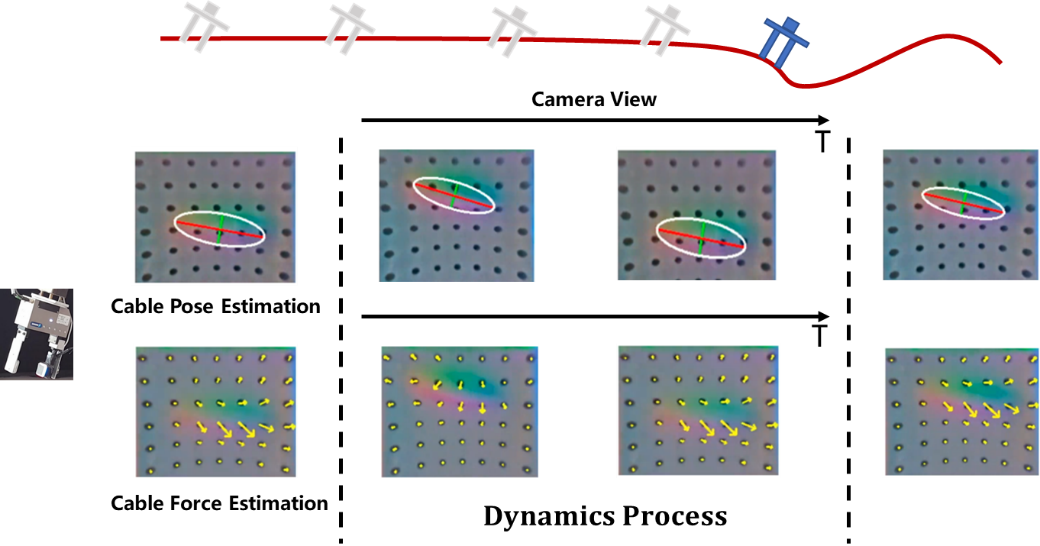}}
\caption{Real-time tactile perception while manipulating DOs (Cable following using GelSight as an example \cite{she2021cable}). The first row shows the gripper grasping a cable. The second row presents the estimated cable poses. The white ellipse represents the estimation of the contact region. The red and green lines denote the first and second principal axes of the contact region, with their lengths proportional to their corresponding eigenvalues. The third row shows a top view of the pulled cable while the gripper registers marker displacements, reflecting the magnitude and direction of the frictional forces.}
\label{tactile}
\end{figure*}

\subsubsection{Tactile devices}
The BioTac tactile sensor is a remarkable device that can acquire sensory modalities mimicking the full range of capabilities of human fingertip \cite{narang2021interpreting}. It has a rigid core that contains all of the sensory electronics and a flexible skin made of low-cost silicone, which enhances the durability and reparability of the design. However, a major limitation of BioTac is that its output signal is subject to complex signal processing and data fusion, increasing its computational effort and latency.

Vision-based tactile sensors are a class of sensors that transform tactile information into visual information by capturing the deformation of a contact surface with a camera. These sensors offer a distinctive benefit of high spatial resolution and have been applied to various robotic manipulation tasks. GelSight, a prominent vision-based tactile sensor, was proposed by Yuan \textit{et al.} in 2017 \cite{yuan2017gelsight}. However, GelSight still has a major drawback of not being able to emulate human tactile sensations such as temperature, humidity, pain, etc.

Furthermore, Sundaram \textit{et al.} \cite{sundaram2019learning} developed a tactile glove that enables passive sensing of various tactile tasks. The glove consists of a knitted glove made of a piezoresistive film with a sensor array of 548 sensors which are connected by a network of conductive wire electrodes. The glove can employ deep convolutional neural networks to process pressure signals for various tactile tasks such as object recognition, weight estimation or tactile pattern exploration.

\subsubsection{DOM applications using tactile sensing}
Although tactile sensing is valuable for manipulation especially when vision is occluded, there are few examples of manipulation tasks that rely solely on tactile sensing. One such task is cable-following or fabric-following. She \textit{et al.} \cite{she2021cable} proposed a method to estimate the pose and the friction force of a cable held by a robot using a tactile sensor based on GelSight technology. The process is shown in the Fig. \ref{tactile}. Hellman \textit{et al.} \cite{hellman2017functional} introduced a real-time tactile perception and decision-making method using BioTac for a tactile-driven contour-tracking task, specifically, closing a sealed bag. Zheng \textit{et al.} \cite{zheng2022autonomous} proposed a method to capture changes in tactile signals during page flipping by Biotac. This study establishes the relationship between tactile signals and a rational page flipping trajectory, enabling a robot to learn to flip pages rationally. Moreover, Building upon the high-resolution tactile glove presented in \cite{sundaram2019learning}, a wide array of interactive tasks involving diverse objects is performed in the system proposed by Zhang \textit{et al.} \cite{zhang2021dynamic}. The tactile model combines a predictive model and a contrastive learning module to estimate the 3-D coordinates of the hand and the object from only the touch data.

\subsection{Discussion of perception}
By touching, humans can simultaneously feel, weigh and manipulate various objects, as well as infer material properties of objects. This is a difficult set of tasks for modern robots. In order to broaden the boundaries of robotic manipulation, multimodal sensory inputs are essential. Vision and touch are two vital senses for robots to perform dexterous manipulation. Relying solely on a single sensory modality to monitor the dynamics of DOs is difficult. For instance, vision can be obstructed, while tactile inputs lack the ability to capture global information for tasks.

In order to realize DOM with multimodal perception, the development of usable tactile perception devices is an important prerequisite. Several tactile sensors have been developed \cite{fishel2012sensing, zhang2022hardware} in recent years. Although tactile sensors have achieved good perception accuracy in reality, simulators for tactile sensors are still rarely investigated and developed due to the difficulty of reconstructing elastomer deformations in optical tactile stimulation. This leads to difficulties in providing a high-fidelity simulation environment for training robots. Multimodal perception holds great promise in the field of robot manipulation, and the research of such tactile simulators is the key to the application of multimodal perception in robot manipulation tasks.

Large-scale and high-quality manipulation datasets are crucial for training robots that can collaborate and perform complex tasks in diverse and unstructured environments. These datasets must span multiple sensing modalities that are well-suited to complex and unstandardized environments. Using the tactile glove developed in \cite{sundaram2019learning}, as well as other wearable sensing devices such as myoelectric sensors and eye-trackers, DelPreto \textit{et al.} \cite{delpreto2022actionsense} developed a multimodal manipulation dataset of a kitchen scene (including DOs). However, this dataset can be further expanded and improved by incorporating more scenarios or activities, as well as by exploring alternative or novel wearable sensors or modalities that can overcome the current limitations of data collection. Producing a dataset of multimodal-aware robot manipulation tasks is another key to the use of multimodal perception in robot manipulation applications.

\begin{table*}
\begin{center}
\caption{Overview of perception approaches of DOs.}
\label{perceptiontable}
\begin{tabular}{c c c c}
\hline
\textbf{Categories} &\textbf{Advantages} &\textbf{Limitations} &\textbf{Representative literature}\\
\hline
\multirow{4}{*}{Visual perception} &\multirow{4}{*}{Acquire global information} 
&\multirow{4}{*}{\makecell[c]{Not robust under occlusion, \\ Affected by environment easily}} &\makecell[c]{DLOs: Tang and Tomizuka \cite{tang2022track} \\ Lv \textit{et al.} \cite{lv2023learning},}\\
& & &\makecell[c]{Cloth-like objects: Chi and Berenson \cite{chi2019occlusion} \\Qian \textit{et al.} \cite{qian2020cloth},} \\
& & &Play-dough: Shi \textit{et al.} \cite{shi2022robocraft}\\
\hline
\multirow{4}{*}{Tactile perception} &\multirow{4}{*}{Acquire local information} & &\makecell[c]{DLOs: Hellman \textit{et al.} \cite{hellman2017functional} \\She \textit{et al.} \cite{she2021cable},}\\
& &\makecell[c]{Tactile information is sparse,\\Requires contact for acquiring}&Page: Zheng \textit{et al.} \cite{zheng2022autonomous},\\
& & &3-D objects: Zaidi \textit{et al.} \cite{zaidi2017model} \\
\hline
\multirow{4}{*}{Multimodel perception} &\multirow{4}{*}{\makecell[c]{Acquire global and\\local information\\at the same time}} &\multirow{4}{*}{\makecell[c]{Hard to represent \\multimodal sensory inputs}} &DLOs: Pecyna \textit{et al.} \cite{pecyna2022visual}\\
& & &Deformable surface: Caccamo \textit{et al.} \cite{caccamo2016active},\\
& & &\makecell[c]{Sponge: Rios \textit{et al.} \cite{arriola2017multimodal} \\anchez \textit{et al.} \cite{sanchez2018online}}\\
\hline
\end{tabular}
\end{center}
\end{table*}

\section{Modeling}
In order to perform DOM, a model that can predict the DO state in the subsequent time steps based on the current state should be available. Such models need to reason the dynamics of DOs. Analytical modeling approaches and data-driven modeling approaches are reviewed in this section. Table. \ref{modelingtable}. provides a glance of the main advantages, limitations and representative literature.

\subsection{Analytical modeling of DOs}
Analytical modeling approaches are under Newton’s second law. In this section, Mass-spring-damper (MSD) systems, Position-based dynamics (PBD) and Continuum mechanics are reviewed.

\begin{figure*}
\centerline{\includegraphics[width=33pc]{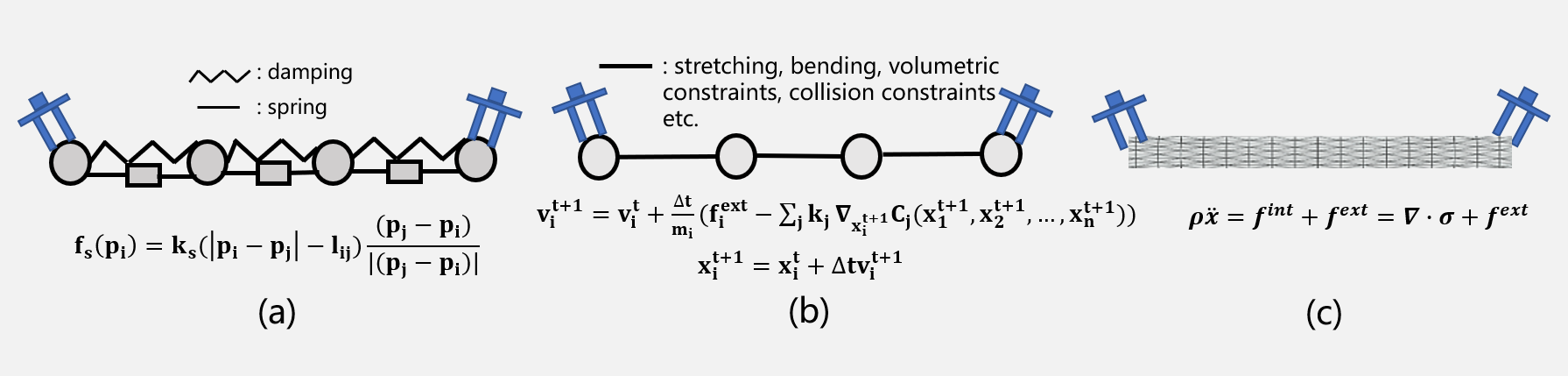}}
\caption{Analytical modeling of DOs (DLOs as an example) (a) A mass-spring-damper system (b) A position-based dynamics system with several constraints (c) Continuum mechanics (FEM as an example)}
\label{analyticalmodel}
\end{figure*}

\subsubsection{Mass-spring-damper systems}
As shown in Fig. \ref{analyticalmodel} (a), MSD systems model deformable materials as network structures. Each particle in the network has a mass attribute, and the connections between particles are modeled by springs and dampers.
For a single particle $i$, the equation of motion is expressed as
\begin{equation}
m_ia_1=f_{ext}(p_i)+f_d(p_i)+f_s(p_i)
\end{equation}
where $m_i$ is the mass of particle $i$, $a_i$ is the acceleration of particle $i$, $f_{ext}$ is the external force applied to particle $i$, $f_d$ is the spring damping force of particle $i$, $f_s$ is the spring force of particle $i$.

MSD systems provide several advantages for simulating DOs, such as intuitiveness, ease of implementation and computational efficiency, which facilitate real-time animations. MSD models are suitable for applications that require a compromise between physical accuracy and computational speed. They have been employed to simulate various kinds of DOs, such as DLO simulation \cite{yu2023coarse} and clothing simulation \cite{kita2011clothes}.

One challenge with MSD systems is they are only suited for simulating small deformations and not complex elastic effects. The system’s behavior does not necessarily converge to the expected outcome by increasing the mesh resolution \cite{muller2008real}. The system is also sensitive to the network topology and the spring parameters. Another challenge arises from lacking physical significance of the spring parameters in relation to the material parameters, which implies that numerous adjustments are necessary to achieve the desired dynamic characteristics. A possible solution to this problem is the application of learning methods and reference schemes \cite{arriola2017multimodal}. MSD models face another challenge in their inability to directly represent volume effects. Teschner \textit{et al.} \cite{teschner2004versatile} addressed this issue by introducing an additional energy equation.

\subsubsection{Position-based dynamics}
PBD shown in Fig. \ref{analyticalmodel} (b). is a particle-based model that incorporates both internal and collision constraints. The constraints include stretching, bending, volumetric constraints, collision constraints, etc. The acceleration of a object are determined by the sum of internal and external forces according to Newton's second law of motion. A time integration method is employed to update the velocities and positions of the object. A set of constraints is solved by an iterative solver loop which utilizes the Gauss-Seidel method, until the convergence criterion is met or the maximum number of iterations is exceeded.

The accelerations of the object are determined by the sum of internal and external forces, according to Newton's second law of motion. A time integration method is employed to update the velocities and positions of the object. A set of constraints is solved by an iterative solver loop which utilizes the Gauss-Seidel method, until the convergence criterion is met or the maximum number of iterations is exceeded.

For a single particle $i$, the equations of motion are expressed as
\begin{equation}
v_i^{t+1}=v_i^t + \frac{\Delta t}{m_i}(f_i^{ext}-\sum_{j}{k_j\nabla_{x_i^{t+1}}C_j}(x_1^{t+1},x_2^{t+1},...,x_n^{t+1}))
\end{equation}
\begin{equation}
x_i^{t+1}=x_i^t+\Delta t v_i^{t+1}
\end{equation}
where $k_j$ denotes the stiffness of the $j^{th}$ constraint $C_j(x_1, x_2, \ldots, x_n)=0, j=1, 2, \ldots.$

PBD is a technique that allows for rapid and fully controllable simulation with imporved stability. The method can easily incorporate various constraints into a system and drive the system by setting boundary conditions. Several popular physics engines, such as PhysX, Bullet and Havok, have implemented PBD in their systems to simulate the dynamics of DOs.

One of the main difficulties in PBD is that the stiffness of constraints is not only determined by the parameters specified by the user, but also by the time step and the number of iterations used by the solver. As the number of iterations increases or the time step decreases, the constraints become more rigid. To address this issue, Macklin \textit{et al.} \cite{macklin2016xpbd} proposed a total Lagrange multiplier method for PBD, which can decouple the constraint solving from these factors.

\subsubsection{Continuum mechanics}
DOs are often modeled as continuous objects shown in Fig. \ref{analyticalmodel} (c). Displacement, strain and stress are the key quantities that significantly impact the understanding of the behaviour characteristics of DOs. The equation of motion for continuum mechanics is described as
\begin{equation}
\rho\ddot{x}=f^{int}+f^{ext}=\nabla\bullet\sigma+f^{ext}
\end{equation}
where $\rho$ denotes density, and force terms are now body forces applied at a unit volume. $\sigma$ represents a stress tensor that is a symmetrical matrix. By applying the gradient operator to the stress tensor, the stress term captures internal elastic effects.

Continuum materials are suitable for modeling the DOs that require high precision in simulation, as they can capture deformation effects more accurately. The model parameters possess a distinct physical significance and can be effectively applied to diverse scenarios with the aid of flexible computing frameworks like FEM. However, the computation and implementation of continuum materials are more complex compared to PBD. Linear FEM finds common usage in robotics situations where deformations remain within a small range. However, unless employing an explicit integration scheme or a hardware-accelerated implementation, nonlinear models typically lack the necessary speed for achieving real-time performance. 

\subsection{Data-driven modeling of DOs}
Data-driven models acquire extensive insights into complex dynamics through direct analysis of data \cite{hu20193, yan2021learning}. Consequently, this method finds utility in planning and control applications. Jacobian-matrix-based models and GNN-based models are reviewed in this article.

\subsubsection{Jacobian-matrix-based model}
The state of DOs can be characterized by a set of features that describe their shape and configuration. Modeling DOs can be regarded as studying the mapping between the state of the end-effectors and the state of the DOs’ features. A Jacobian matrix can be used to locally approximate this mapping. It is described as
\begin{equation}
\dot{x}=\ J(x)v
\end{equation}
where \textit{v} is the velocity of end-effectors, $\dot{x}$ is the velocity of DOs’ features, $\ J(\cdot)$ is a Jacobian matrix which is fully determined by the DOs’ current state \textit{x}.

\begin{figure*}
\centerline{\includegraphics[width=28pc]{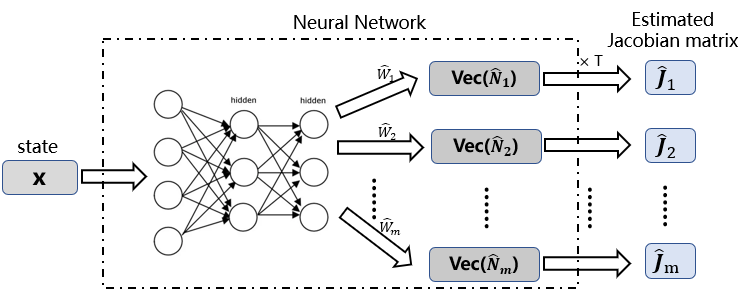}}
\caption{Architecture of a model to output the estimated Jacobian matrixes \cite{yu2022global}. A Jacobian matrix is fully determined by the DOs’ current state x.}
\label{Jacobianmodel}
\end{figure*}

These models have a linear structure and can be computed in real time with a low data volume. Compared to directly modeling with deep neural networks (DNNs), modeling approaches based on a Jacobian matrix can make more efficient use of data.

Some researchers have employed exclusively online techniques to estimate the Jacobian matrix to build a local linear deformation model of DOs \cite{navarro2013model, navarro2016automatic, zhu2018dual, jin2019robust, lagneau2020automatic, zhu2021vision}. Due to their online execution, these models can be readily applied to a wide range of new objects. However, the utilization of a limited amount of local data in these estimated models compromises their accuracy, limiting their effectiveness to local configurations. Consequently, this method can only handle tasks with small and localized deformations. For globally modeling and control of DOs, Yu \textit{et al.} \cite{yu2022global} developed a method to efficiently learn the global deformation model of DOs by integrating offline learning and online adaptation. They employed a radial basis function neural network (RBFN) to approximate the nonlinear mapping from the current state of the DO to the current linear deformation model, as illustrated in Fig. \ref{Jacobianmodel}. The RBFN was trained offline with randomly sampled data, then transferred to the online phase as an initial estimate, and finally updated to adapt to the DOs during real-world manipulation. Their framework achieves global modeling of larger deformation using the Jacobian-matrix-based approach, although the term “large” is relative to the existing Jacobian-matrix-based studies.

\subsubsection{GNN-based model}
In addition to Jacobian-matrix-based modeling approaches, global models of DOs can be approximated with DNNs. DNN-based approaches have a stronger representation power than Jacobian-matrix-based approaches, which enhances their accuracy and robustness \cite{valencia2020combining}. Furthermore, they can integrate physics models and infer how objects interact \cite{battaglia2016interaction}. These models can cope with very complex systems and perform well in a broader range of situations, overcoming (to some extent) the drawbacks of linear models that are constrained by local conditions. However, such approach requires substantial datasets that may be unavailable in certain scenarios.

\begin{figure*}
\centerline{\includegraphics[width=28pc]{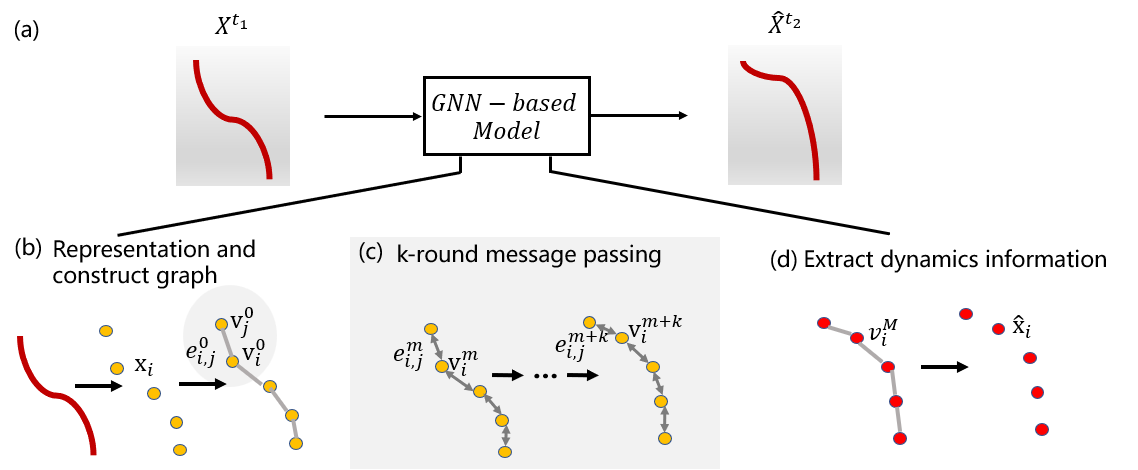}}
\caption{One-step prediction through GNN-based Model (Message-passing GNN type for manipulating a DLO as an example \cite{li2019propagation, sanchez2020learning}) (The model is IN, when k equals 1.) (a) The GNN-based model predicts future states represented as particles using its learned dynamics model. (b) It constructs a latent graph from the input state. (c) It performs k rounds of learned message-passing over the latent graphs. (d) It extracts the dynamics information from the final latent graph.}
\label{GNNmodel}
\end{figure*}

Characteristics of DOs are complex, resulting in capturing the behavior of various DOs is challenging for those generic network structure such as MLP, CNN and LSTM. Research \cite{xia2021graph} on GNNs has brought a new structure for modeling DOs. Battaglia \textit{et al.} \cite{battaglia2016interaction} proposed Interaction Network (IN), a general-purpose, learnable physics engine that enables object- and relation-centered reasoning about physics. IN represents the particles of an object as nodes of a graph, and the interactions between pairs of nodes as edges of the graph. IN uses an object function and a relation function to model objects and their relations in a combinatorial manner. IN can learn pairwise local interactions between different parts of a DO. The process of prediction for IN is shown in Fig. \ref{GNNmodel}.

Assuming two objects and one directed relationship, e.g., a fixed object connected to a movable mass by a spring, the first object (the sender $o_1$) affects the second (the receiver $o_2$) via their interaction. The effect of this interaction $e_{t+1}$ can be estimated by a relation-centric function $f_R$. The $f_R$ takes $o_1$, $o_2$ and attributes of their relationship $r$, such as the spring constant, as input. The $f_O$ is adapted to take both $e_{t+1}$ and the receiver’s current state $o_{1,t}$ as input, allowing the interaction to influence its subsequent state $o_{2,t+1}$. They are expressed as
\begin{equation}
e_{t+1}=f_R(o_{1,t},o_{2,t},r)
\end{equation}
\begin{equation}
o_{2,t+1}=f_O(o_{2,t},e_{t+1})
\end{equation}

One of the limitations of IN is its failure to propagate the effect of the interaction throughout the entire object. To address this problem, Some studies \cite{li2019propagation, li2018learning, yang2021learning} have proposed different methods. Li \textit{et al.} \cite{li2019propagation} introduced a propagation network called PropNet to address the drawback of IN's inability to propagate interactions across the entire object by doing multi-step information transfer. Specifically, the shared information is first encoded and the propagation steps are reused many times. Propagation networks need to be repeated many times to deal with long distance dependencies, which is both inefficient and difficult to train. Therefore, inspired by Mrowca  \textit{et al.} \cite{mrowca2018flexible}, DPI-Net proposed by Li \textit{et al. } \cite{li2018learning} adds an additional level to efficiently propagate long-distance effects between particles. Yang \textit{et al.} \cite{yang2021learning} proposed BiINLSTM, which combines the IN dynamics learning method with a recursive model (e.g., LSTM), to capture the interaction between neighboring particles in a DLO. Furthermore, it utilizes the recursive model to propagate the interaction effect along the entire length of the DLO. Unlike the original IN approach, the BiINLSTM model enables the transmission of local interactions to every particle in the DLO within one time step, eliminating the need for repeated propagation as described in \cite{li2019propagation}. But this method is still only verified to work in DLOs.

GNN-based approaches are a powerful way of representing structure information, but they have some limitations. The expression of concepts like recursion, control flow, and conditional iteration through graphs can pose challenges unless specific assumptions are made, such as employing abstract syntax trees. Programs and more “computer-like” processing can have more expressive power and flexibility for these concepts, and some researchers think they are an important part of human cognition (Tenenbaum \textit{et al.} \cite{tenenbaum2011grow}; Lake \textit{et al.} \cite{lake2015human}; Goodman \textit{et al.} \cite{goodman2014concepts}).

Data-driven approaches rely on simulators to generate data. As a result, the gap between simulation and real-world scenarios remains a major obstacle for applying the model to robotic manipulation tasks in the real world. Several studies \cite{wang2022offline, zhang2022learning, shi2022robocraft} have proposed different methods to address the issue of sim-to-real gaps. Wang \textit{et al.} \cite{wang2022offline} employed a GNN to learn a dynamics model from synthetic data. Then, an online linear residual model is learned to reduce the gap between simulation and reality. The work proposed by Zhang and Demiris \cite{zhang2022learning} compares real and simulated pairs of object observations to learn their physical similarities, which allows the simulator to better adapt to real-world objects’ physics. The study introduced by shi \textit{et al.} \cite{shi2022robocraft} obtains data directly from real observations and uses a GNN to learn particle-based dynamics models to perform deformation operations on Play-Doh. All three methods mentioned are effective (to some extent) for the problem of the sim-real gap.

\subsection{Discussion of modeling}
Obtaining the exact deformation model of DOs is a challenging task since they are difficult to calculate theoretically. Analytical modeling approaches are not capable of accurately modeling DOs due to their infinite state dimensions and the difficulty in acquiring DOs’ parameters in the real world. Data-driven approaches can learn a fairly accurate model from data without understanding the complex physical dynamics, given the availability of sufficient high-quality data. The deformation models may differ considerably across different DOs owing to their diverse lengths, thicknesses, materials, etc. It is impractical to spend a large amount of time to collect new data for every DOM task.

The predictive power of both analytical and learned models is limited. These models are only effective for certain types of tasks. Therefore, researchers  have proposed evaluation methods \cite{mcconachie2020learning, power2021keep, mitrano2021learning} to assess the reliability of models during a manipulation task. In situations where the model exhibits unreliability, replanning or recovery strategies should be considered.

\begin{table*}
\begin{center}
\caption{Overview of modeling approaches of DOs.}
\label{modelingtable}
\begin{tabular}{c c c c}
\hline
\textbf{Categories} &\textbf{Advantages} &\textbf{Limitations} &\textbf{\makecell[c]{Representative literature}}\\
\hline
\multicolumn{4}{l}{\textbf{Analytical approaches}}\\
\hline
\multirow{6}{*}{MSD} &\makecell[c]{\\Fast,} & &\makecell[c]{DLOs: Bergou \textit{et al.} \cite{bergou2008discrete}\\ Lv \textit{et al.} \cite{lv2017physically},} \\
&\makecell[c]{Computationally\\efficient,}&\makecell[c]{Inaccurate for\\large deformation,} & \makecell[c]{Cloth-like objects: Kita \textit{et al.} \cite{kita2011clothes}\\ Goldenthal \textit{et al.} \cite{goldenthal2007efficient},} \\
&\makecell[c]{Easy to\\implement} &\makecell[c]{Require accurate\\physical parameters} &\makecell[c]{Tissue: Kuhnapfel \textit{et al.} \cite{kuhnapfel2000endoscopic} \\Mollemans \textit{et al.} \cite{mollemans2003tetrahedral}} \\
\hline
\multirow{6}{*}{PBD} &\makecell[c]{\\Fast,} & &  \\
&\makecell[c]{Fully\\controlled,} &Visual fidelity only, &\makecell[c]{Cloth-like objects: Macklin \textit{et al.} \cite{macklin2014unified}\\Macklin \textit{et al.} \cite{macklin2016xpbd},}\\
&\makecell[c]{Improved\\stability} &\makecell[c]{Lacking physical\\interpretability} &Page: Zollhöfer \textit{et al.} \cite{zollhofer2014real}\\
\hline
\multirow{6}{*}{Continuum mechanics}& & &\makecell[c]{DLOs: Koessler \textit{et al.} \cite{koessler2021efficient} \\Duenser \textit{et al.} \cite{duenser2018interactive},}\\
& &Heavy computational&Cloth-like objects: Cusumano-Towner \textit{et al.} \cite{cusumano2011bringing},\\
&Accurate, &requirements, &Muscles: Chen \textit{et al.} \cite{chen1992pump},\\
&Physical &Difficult to achieve &Ring-like objects: Yoshida \textit{et al.} \cite{yoshida2015simulation},\\
&interpretation &real-time performance &Silicone: Kinio \textit{et al.} \cite{kinio2012comparative}\\
\hline
\multicolumn{4}{l}{\textbf{Data-driven approaches}}\\
\hline
\multirow{6}{*}{Jacobian-matrix-based} & & &\\
& & &\\
&\makecell[c]{Require a small\\amount of data,} &\makecell[c]{Only model\\small deformation}  &\makecell[c]{DLOs:  Lagneau \textit{et al.} \cite{lagneau2020automatic}\\Yu \textit{et al.} \cite{yu2022global},}\\
&\makecell[c]{computationally\\efficient}& &\makecell[c]{3-D objects: Navarro-Alarcón \textit{et al.} \cite{navarro2013model}\\Navarro-Alarcón \textit{et al.}  \cite{navarro2016automatic}}\\
\hline
\multirow{6}{*}{GNN-based} & & &\makecell[c]{DLOs: Yang \textit{et al.} \cite{yang2021learning}\\Li \textit{et al.} \cite{li2018learning},}\\
&\makecell[c]{Model complex\\dynamics accurately,} &\makecell[c]{Require a large\\ amount of data,} &\makecell[c]{Cloth-like objects: Li \textit{et al.} \cite{li2020causal}\\ Huang \textit{et al.} \cite{huang2022mesh},} \\
&Flexible &\makecell[c]{Lacking\\physical interpretability}&\makecell[c]{Dough-like objects: Shi \textit{et al.} \cite{shi2022robocraft}\\Shi \textit{et al.} \cite{shi2023robocook}}\\
\hline
\end{tabular}
\end{center}
\end{table*}

\section{Manipulation}
The objective of manipulating DOs is to determine the optimal force or motion at driving points to achieve the given task goal. We firstly review works on analytical planning and control for DOM. Then, we mainly focus on reviewing learning methods such as RL and IL. Table. \ref{manipulationtable}. presents a comprehensive overview of manipulation, and summarizes representative literature.

\subsection{Planning for DOM}
Planning is the process of finding an optimal sequence of configurations for a robot or an object to achieve a desired goal. Let us consider a task such as bending a rope into a certain shape.  A common method to solve this problem is to formulate it as an optimization problem
\begin{equation}
x_{0:T}^\ast,u_{0:T-1}^\ast= \mathop{\arg \min} \limits_{x_{0:T},u_{0:T-1}}{\mathcal{J}_{u_{0:T-1}}(x_{0:T})}
\end{equation}
\begin{equation}
x_{t+1}=SystemDynamics(x_t,u_t)
\end{equation}
where $x_(0:T)$ denotes the action of manipulating DOs, and $\mathcal{J}(\cdot)$ denotes the total cost incurred in executing the planned trajectory $x_{0:T}$, $u_{0:T-1}$ in the time range $T$.

\begin{figure*}
\centerline{\includegraphics[width=28pc]{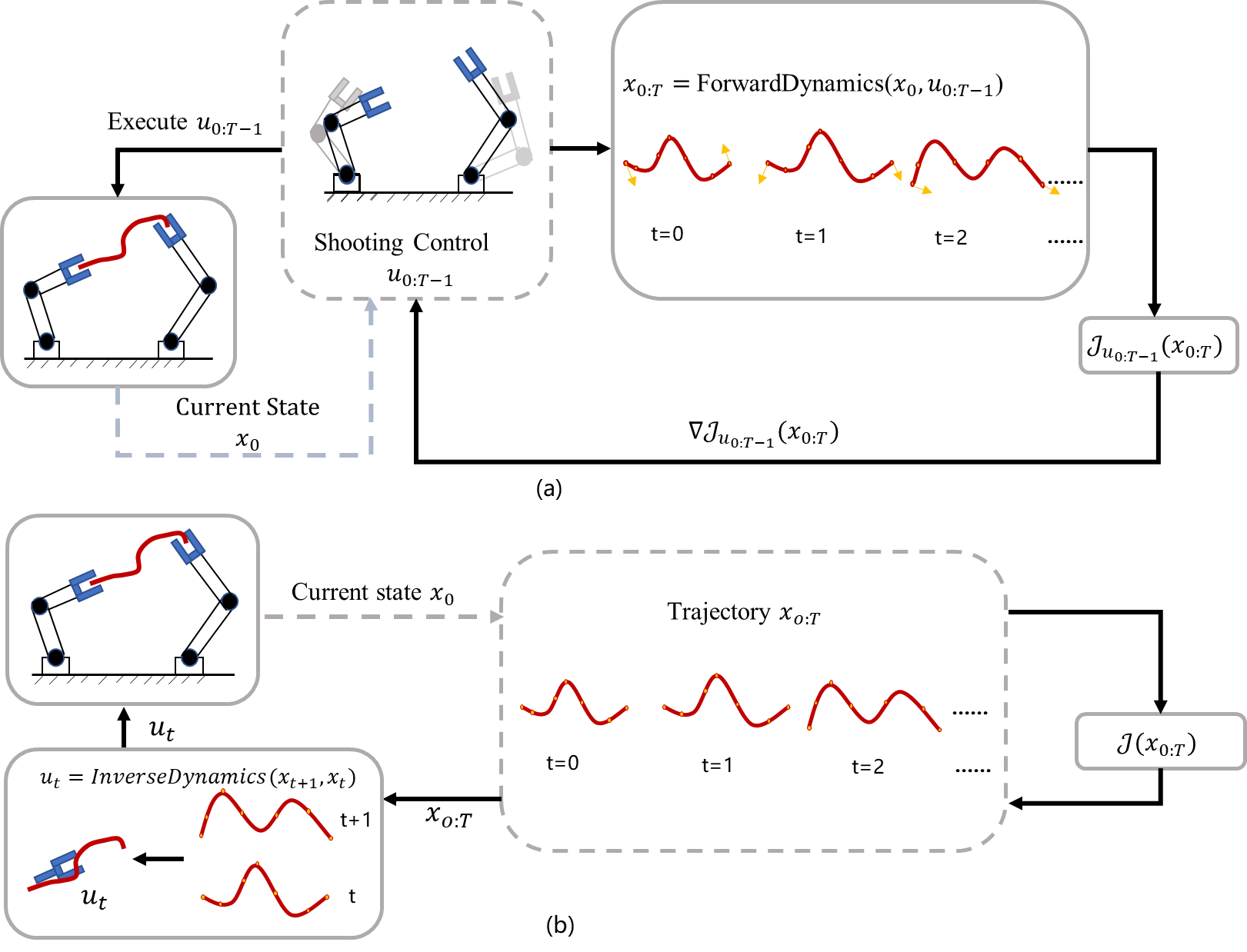}}
\caption{Planning manipulation of DOs. (DLOs as an example) (a) Action shooting: the decision variables are in the action and control space (dashed rectangle). (b) State trajectory exploration: the decision variables are in the state space (dashed rectangle), and actions are derived from state trajectories and inverse dynamics. }
\label{Planning}
\end{figure*}

\subsubsection{Shooting in the action space}
A possible solution to Eq. 9 is to apply a sample action sequence and then update it based on the computed cost of the predicted trajectories. Fig \ref{Planning} (a) illustrates the process of shooting in the action space. In general, the action space has a much lower dimensionality than the object state space, which facilitates the search in the action space. Precise dynamics modeling and efficient backward gradient evaluation are essential for this type of planning method.

Huang \textit{et al.} \cite{huang2023learning} proposed a method to perform shape control manipulation on DLOs using a dual-arm robotic system by searching actions in the action space. Shi \textit{et al.} \cite{shi2022robocraft} proposed a method to   deform Play-Doh into various letter shapes with capabilities comparable to humans using such planning method. Other studies \cite{li2015folding, lin2015picking, zaidi2017model} also focus on generating manipulation actions relying on action-space shooting.

\subsubsection{Searching state trajectories}
One possible approach to plan manipulation actions is to explore in the object state space, e.g., finding a low-cost path of $x_t$, and then generating the corresponding actions for each temporal transition. The process of searching state trajectories is illustrated in Fig. \ref{Planning} (b). The challenge here is to efficiently search or sample configurations that are dynamically feasible in a large state space. Therefore, related research focuses on defining state representation or transition models in an efficient way.

Sintov \textit{et al.} \cite{sintov2020motion} demonstrated that the configuration space of elastic rods, which represents the set of all possible positions and orientations of the rods, is a smooth manifold with finite dimension. They also proposed a graph-based parameterization of this manifold and a sampling-based algorithm to compute a collision-free and kinematically feasible path between any two stable rod configurations. The research proposed by Lui and Saxena \cite{lui2013tangled} improves the efficiency of planning for DLO shaping tasks by using a simple yet effective DLO energy model to calculate a coarse path that guaranteed the achievability of the task. They also designed a local controller that follows this path and adjusts it with closed-loop feedback to compensate for planning errors and achieve task accuracy. Other studies \cite{bretl2014quasi, shah2016towards, mcconachie2020learning} also focus on defining state representation or a transition model efficiently to search state trajectories.

\subsection{Control strategies for DOM}
Control aims at designing inputs for robots to execute desired motions. Closed-loop control which utilizes sensory feedback can cope with various sources of uncertainty in DOM. However, due to the complex dynamics of DOs, it is not feasible to synthesize a global control strategy with theoretical guarantees. Therefore, local controllers are employed to exert force or velocity at designated operating points, typically following
\begin{equation}
u_t=u(\emptyset(x_t)-\emptyset(x_g),\theta)
\end{equation}
where $x_t$ is the current state, $x_g$ is the goal state and $\phi(\cdot)$ is the feature space.

\begin{figure*}
\centerline{\includegraphics[width=28pc]{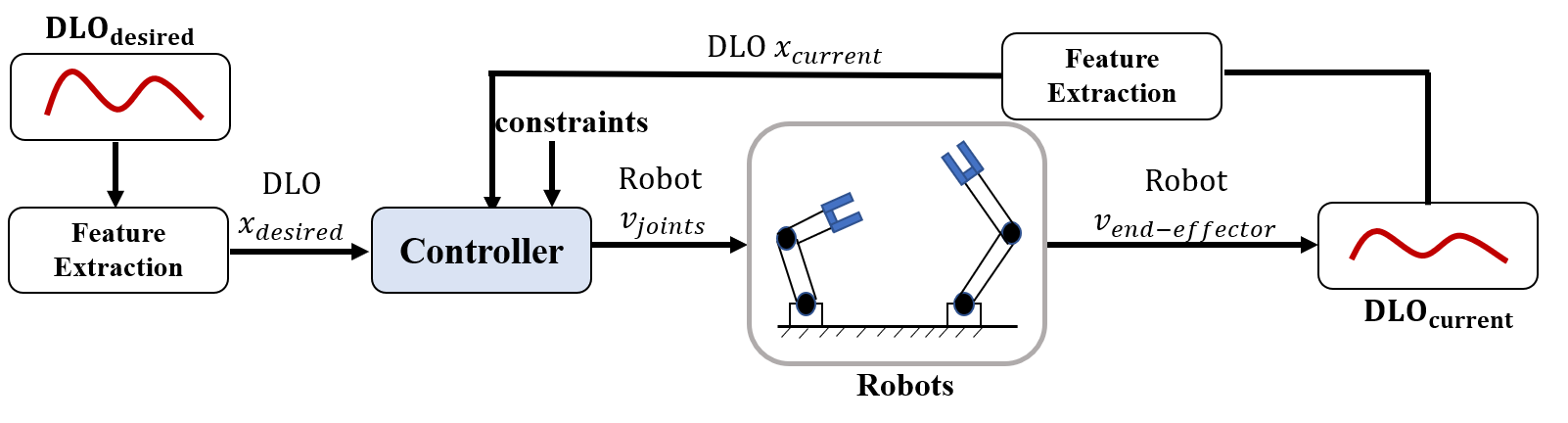}}
\caption{A simple closed-loop control framework  based on visual features (DLOs manipulation as an example).}
\label{control}
\end{figure*}

Control minimizes the mismatch between the current state $x_i$ and the target state $x_g$, possibly in a feature space $\phi(\cdot)$. Due to the real-time constraint, $u(\cdot)$ is usually computed using simple methods, such as a linear function between operating and feature points \cite{berenson2013manipulation, das2010planning, navarro2017fourier} or a numerical optimization procedure with a single-step horizon \cite{ruan2018accounting, ficuciello2018fem}. A simple control framework for DOM is shown in Fig. \ref{control}.

Some examples use pure visual-servoing frameworks to regulate velocities, e.g., arranging cloth \cite{kruse2015collaborative, ruan2018accounting}, manipulating DLOs \cite{li2018vision, bretl2014quasi, roussel2015manipulation, lv2022dynamic, koessler2021efficient}. On the other hand, tactile features are used to adjust the manipulation actions, e.g., cable manipulation \cite{she2021cable}.

\subsection{Learning-based approach}
A variety of DOM tasks can be achieved by learning-based approaches. This review focuses on RL and IL. Learning-based methods provide the advantage of eliminating the need for DOs' dynamical model which is difficult to be acquired for many DOs.

\subsubsection{Reinforcement learning for DOM}
\begin{figure*}
\centerline{\includegraphics[width=28pc]{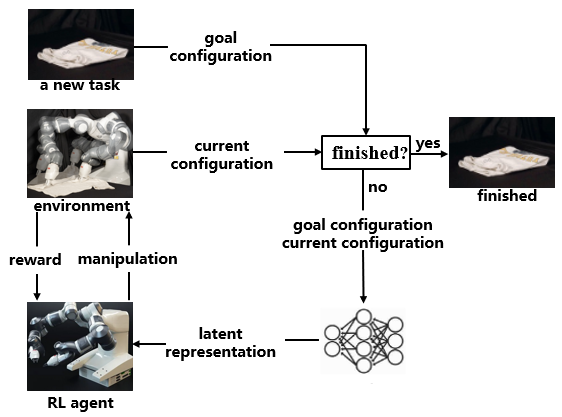}}
\caption{RL pipeline (Cloth folding as an example). First, the configurations of the DO are encoded into a latent space. Then, the RL agent outputs manipulation actions that are mapped from the current and goal latent representation to interact with the environment continuously until the task is completed.}
\label{RL}
\end{figure*}

RL is a paradigm for learning optimal control policies from trial-and-error interactions with an environment. The learning objective is to maximize the cumulative rewards obtained during the interactions. Fig. \ref{RL}. shows a pipeline of RL.

RL aims to find optimal policy $\pi$ by maximizing the expected future accumulative reward. It can be expressed as
\begin{equation}
\pi^\ast=\mathop{\arg \max} \limits_{\pi \in \Pi} \mathbb{E}_{\tau \in d^\pi}({\sum_{t=1}^{T}{\gamma^{t-1}r_t}})
\end{equation}
where $\tau$ denotes a trajectory from the distribution $d^\pi$ of trajectories generated by the MDP or policy $\pi$. $r_t$ denotes the reward collected at step $t$. $\gamma$ represents the discount factor ranging from 0 to 1 to regulate the significance of a reward with the increment of its collected future step.

Generally, an explicit dynamics model of DOs is not required in model-free RL. Despite this, RL faces its own set of challenges. One such challenge is the exploration process is often sample inefficient, especially for DOs with infinite dimensions. Another challenge arises from the inherent danger of conducting tens of thousands of trial-and-error interactions in real environments. The development of simulators has facilitated training in simulated environments. However, the primary challenge lies in transferring learned strategies to real robots.

Embedding structures or reducing the dimensionality of features can be performed in the strategy design to improve RL efficiency. Wu et al. \cite{wu2020learning} suggested an iterative pick-and-place action space that takes into account the conditional relationships between picking and placing actions on DOs. This approach employs explicit structural encoding to facilitate learning an DOs’ organizing skill. Colom{\'e} \textit{et al.} \cite{colome2018dimensionality} investigated the simultaneous learning of a DMP-characterized robot motion and its underlying joint couplings using linear dimensionality reduction. This approach provides valuable qualitative information, resulting in a concise and intuitive algebraic description of the motion. Moreover, the proposed method efficiently reduces the number of parameters to be explored for RL algorithms.

Model-free RL methods, which are widely used for deep RL algorithms, suffer from high data demand that hinders their applicability to real-world robotics problems. In contrast, model-based RL approaches are expected to achieve much better data efficiency than model-free RL methods, provided that the model is accessible \cite{wang2019benchmarking}. A novel model-based RL system that incorporates sensing information, named SAM-RL, was developed by Lv \textit{et al.} \cite{lv2022sam}. SAM-RL utilizes differential physics-based simulation and rendering to dynamically update the model. This process involves comparing the rendered image with the real original image, leading to the efficient generation of policies. The system demonstrates remarkable performance in a needle-threading task, achieving a substantial reduction in training time and a significant increase in success rate. In addition, Yang \textit{et al.} \cite{yang2022online} employed a framework of model-based RL that interleaves learning with exploration to solve a 3-D shape control task of DLOs. The proposed method demonstrates superior cable shape control performance while utilizing significantly fewer interaction samples than state-of-the-art model-free RL methods.

In order to reduce the cost of strategy iteration, initializing RL by manual demonstration has become a feasible way. Zheng \textit{et al.} \cite{zheng2022autonomous} initialized a page-flipping trajectory manually in a real robot. The page-flipping trajectory is then further learned by using feedback from tactile sensors as a reward, allowing a real robot to learn how to turn a book in just 200 interactions of trial and error. Moreover, the work proposed by Tsurumine \textit{et al.} \cite{tsurumine2019deep} has demonstrated the capability to learn laundry folding with mere 80 samples for strategy initialization and 100 samples for RL.

The emergence of simulators has enabled the execution of learning iterations in simulated environments. However, when transferring the learned policies to the real world, there is often a discrepancy between simulation and reality. Domain randomization (DR) is a technique that aims to mitigate this discrepancy between visual simulation and reality by randomly augmenting the visual parameters of the simulation, such as texture and illumination. This approach enables the strategy to learn generic and task-relevant visual features. Despite the success of DR methods in achieving cross-domain generalization \cite{peng2018sim}, they still face challenges in terms of task-specificity and adaptability. The work proposed by Matas \textit{et al.} \cite{matas2018sim} uses a visual RL pipeline based on pixel-level domain adaptation to bridge the gap between simulation and reality, and demonstrates its effectiveness in a tissue retraction medical manipulation task. Besides, learning a residual strategy to bridge the gap between reality and simulation is another significant approach. For instance, the work proposed by Lv \textit{et al.} \cite{lv2022sam} successfully performs the task of threading a needle in the real world by learning a residual strategy. This shows that learning a residual strategy can help alleviate the sim-real gap.

\subsubsection{Imitation learning for DOM}
\begin{figure*}
\centerline{\includegraphics[width=28pc]{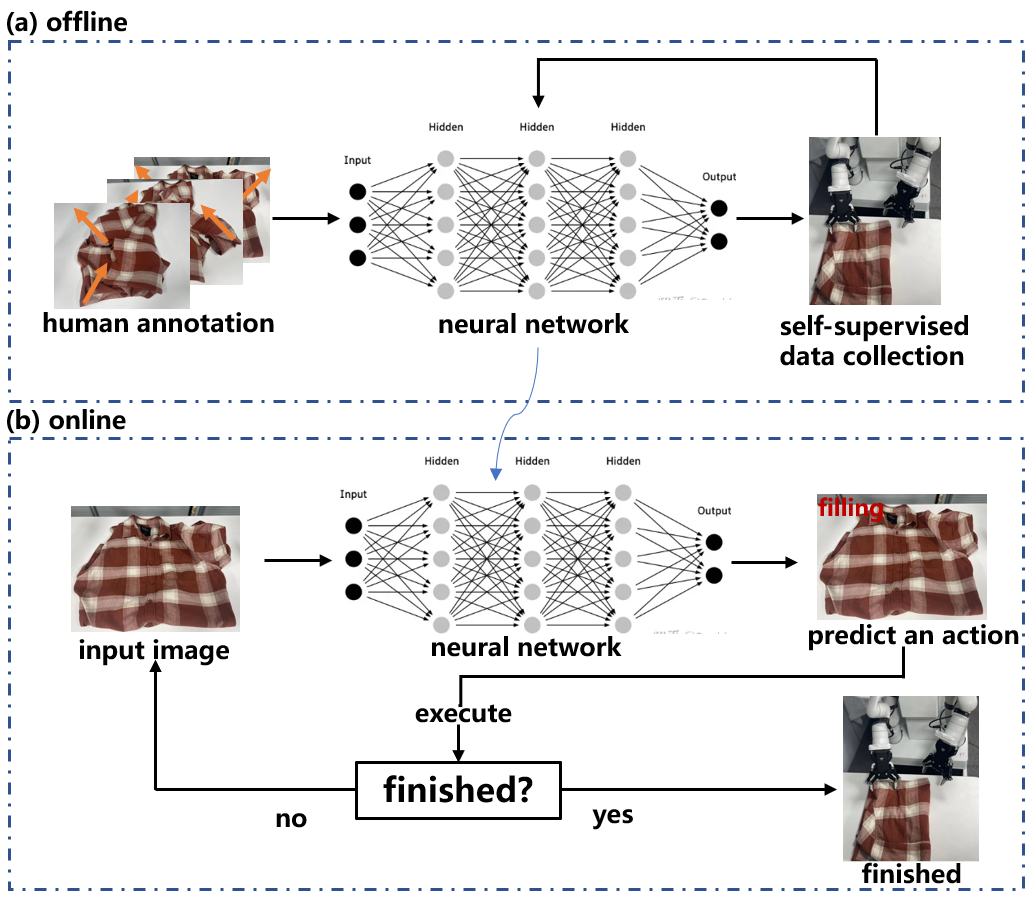}}
\caption{IL Pipeline (Cloth folding as an example \cite{avigal2022speedfolding}). (a) Offline: Input images are manually annotated with primitives and gripper poses. A neural network is trained on the data and iteratively used for self-supervised data collection. (b) Online: The neural network is used to predict a primitive and a pair of poses given an input image, and execute it on the robot. If the resulting garment configuration is classified as finished, the robot will stop. Otherwise, it will repeat the process.}
\label{IL}
\end{figure*}

IL is a control technique that leverages data from expert demonstrations to learn a policy that maps raw inputs to actions. IL has been successfully applied to various domains of manipulation \cite{osa2018algorithmic, rivera2022visual, wu2018action}. Fig. \ref{IL}. shows a pipeline of IL. The objective of IL is to acquire a policy that can accurately imitate the behavior of an expert demonstrator. This method can be mathematically expressed as 
\begin{equation}
\pi^\ast=\mathop{\arg \min} \limits_{\pi \in \Pi} \mathbb{D}(p(\pi^{demo})||p(\pi))
\end{equation}
where $\mathbb{D}$ is the divergence metric that quantifies the dissimilarity between the two distributions. The operands in the divergence can be the distribution of state–action pairs, trajectory feature expectation or the distribution of trajectory features.

A key distinction between IL and RL is that IL does not require a reward function, thus eliminating the need for reward engineering which is difficult to achieve in many robotic applications. However, this method has some limitations, such as the need for extensive demonstrations and poor generalization.

Early work on IL focuses on extracting primitive motions and enabling robots to learn primitive motions for DOM through kinesthetic teaching. For example, a method for aerial knot tying by a two-armed, multi-fingered robot was proposed by Kudoh \textit{et al.} \cite{kudoh2015air}. Knot tying tasks are decomposed into six primitive motions, namely grasping, releasing, double grasping, wrapping, twisting, and sliding. The robot learns these six primitive motions through human demonstrations and combines them to perform various knot tying tasks. In addition to aerial knotting, this method and its variants realize rope tying \cite{lee2015learning}, fabric folding \cite{schulman2016learning} and stitching \cite{schulman2013case}. A further extension is to incorporate task parameters to models for enhancing the generality of models. For example, the dressing motion can be adapted to different arm postures by using task-parameterized models \cite{pignat2017learning}. Recent research has made it possible to plan actions based on current observations. This is achieved by collecting and annotating a large amount of demonstration data and extracting features \cite{yang2016repeatable, cherubini2018towards, seita2020deep, ha2022flingbot, avigal2022speedfolding}. For example, a series of two-arm action primitives for folding clothes are first parameterized \cite{avigal2022speedfolding}. After learning from 4,300 human-labeled actions, the system can predict pairs of jaw-clamping poses based on visual information. It has been able to fold randomly placed garments with a 93$\%$ success rate in an average of 120 seconds.

IL methods require retraining to transfer learned skills from known instances to similar/new instances. In the case of a hat-wearing task, the hat-wearing step needs to be repeated thousands of times during the training phase in both simulated and real-world environments. However, to process a new hat, we need collect data new data and retrain the model for the new hat, due to the variations in its shape and deformation. Consequently, these steps impose significant computational and temporal costs, which hinders the feasibility of this approach in real-world scenarios. Recently, several studies \cite{simeonov2022neural, wen2022you} have explored IL for category-level manipulation for rigid objects. However, these methods cannot be applied to DOs. Unlike rigid objects whose poses can be fully represented by low-dimensional vectors, DOs have an infinite configuration space that is prone to severe self-occlusion. These features pose a challenge for skill generalization across different DOs. Recently, Ren \textit{et al.} \cite{ren2023autonomous} proposed a novel framework for manipulating deformable 3-D objects at the category level, which enables the transfer of learned skills to new instances of similar objects with only a single demonstration. Specifically, the framework consists of two modules. The first module, Nocs State Transition (NST), converts the observed target point cloud to a predefined uniform pose state (i.e., Nocs state), which serves as the foundation for category-level skill learning. The second module, Neuro-Spatial Encoding (NSE), adapts the learned skills to new instances by encoding category-level spatial information to infer the optimal grasping point without retraining.

\subsection{Discussion of manipulation}
Currently, significant progress in short-horizon tasks for DOM has been made. However, real-world manipulation tasks often require a sequence of subtasks that have distinct characteristics and objectives. These long-horizon and complex tasks demand dexterous hands or multiple tools to perform different subtasks. Compositional generalization for subskills is regarded as a crucial challenge for long-horizon tasks.

Dexterous hands possess adaptability and versatility, enabling smooth transitions between various functional modes without relying on external tools. However, the high-dimensional action space of dexterous hands poses challenges. Some research has shown dexterous hands can be utilized for manipulation of rigid objects. Chen \textit{et al.} \cite{chen2023sequential} presented a comprehensive system that leverages RL to chain multiple dexterous policies for achieving complex task goals. However, their system only deals with rigid object manipulation. The application of dexterous hands to long-horizon tasks for DOM remains an area that requires further investigation.

In addition to dexterous hands, the usage of diverse tools is also a promising approach to tackle long-horizon tasks. Human beings demonstrate remarkable proficiency in performing complex and long-horizon manipulation tasks via flexible tool use. However, robotic tool use is still constrained by the difficulties in understanding tool-object interactions. RoboCook proposed by Shi \textit{et al.} \cite{shi2023robocook}, manipulates elasto-plastic objects such as dough for long-horizon tasks using multiple tools. However, failures such as dough sticking to the tool often occur. The utilization of multiple tools to accomplish long-horizon tasks warrants further investigation.

Compositional generalization for subskills is a significant challenge for long-horizon tasks. Planning skill sequences with a large search space is inherently difficult, but the motion planning community has achieved remarkable advances in producing a plan skeleton \cite{garrett2017sample, kim2020learning}. For instance, Garrett \textit{et al.} \cite{garrett2017sample} suggested two approaches. The first approach involves alternating between searching for the skill sequence and performing lower-level optimization. The second approach employs lazy placeholders for some skills. Subsequent studies have also investigated the use of pre-trained LLMs for discovering skill sequences \cite{brohan2023can, huang2022language}. Compositional generalization for subskills is a challenging problem that requires further study.

\begin{table*}
\begin{center}
\caption{Summary of main literature on manipulation planning, control, and learning.}
\label{manipulationtable}
\begin{tabular}{c c c c c c c c c c}
\hline
\textbf{Year}&\textbf{\makecell[c]{Representative literature}}&\multicolumn{3}{c}{\textbf{Type of objects}}&\multicolumn{2}{c}{\textbf{Peception}}&\multicolumn{3}{c}{\textbf{Modeling}}\\
\hline
& &\textbf{\makecell[c]{1-D\\objects}} &\textbf{\makecell[c]{2-D\\objects}} &\textbf{\makecell[c]{3-D\\objects}} &\textbf{Visual} &\textbf{Tactile} & \textbf{\makecell[c]{Analytical\\approaches}} &\textbf{\makecell[c]{Data-driven\\approaches}} &\textbf{None} \\
\hline
\multicolumn{10}{l}{\textbf{Analytical planning and control}}\\
\hline
2014 &Bretl and McCarthy \cite{bretl2014quasi} &\checkmark & & & \checkmark & &\checkmark & & \\
2015 &Li et al. \cite{li2015folding} & &\checkmark & & \checkmark & &\checkmark & & \\
2015 &Li et al. \cite{lin2015picking} & & &\checkmark & \checkmark & &\checkmark & & \\
2017 &Zaidi et al. \cite{zaidi2017model} & & &\checkmark & &\checkmark &\checkmark & & \\
2018 &Navarro-Alarcon and Liu \cite{navarro2017fourier} & &\checkmark & &\checkmark & &\checkmark & & \\
2018 &Li et al. \cite{li2018vision} &\checkmark & & &\checkmark & &\checkmark & & \\
2020 &Sintov et al. \cite{sintov2020motion} &\checkmark & & &\checkmark & &\checkmark & & \\
2020 &McConachie et al. \cite{mcconachie2020learning} &\checkmark & & &\checkmark & &\checkmark & & \\
2021 &Koessler et al. \cite{koessler2021efficient} &\checkmark & & &\checkmark & &\checkmark & & \\
2021 &She et al. \cite{she2021cable} &\checkmark & & & &\checkmark &\checkmark & & \\
2022 &Shi et al. \cite{shi2022robocraft} & & &\checkmark &\checkmark & & &\checkmark & \\
2022 &Lv et al. \cite{lv2022sam} &\checkmark & & &\checkmark & &\checkmark & & \\
2023 &Huang et al. \cite{huang2023learning} &\checkmark & & &\checkmark & & &\checkmark & \\
\hline
\multicolumn{10}{l}{\textbf{Reinforcement learning}}\\
\hline
2018 &Peng et al. \cite{peng2018sim} & & &\checkmark &\checkmark & & & &\checkmark \\
2018 &Colomé and Torras \cite{colome2018dimensionality} & &\checkmark & &\checkmark & & & &\checkmark \\
2018 &Matas et al. \cite{matas2018sim} & &\checkmark & &\checkmark & & & &\checkmark \\
2019 &Tsurumine et al. \cite{tsurumine2019deep} & &\checkmark & &\checkmark & & & &\checkmark \\
2020 &Wu et al. \cite{wu2020learning} & &\checkmark & &\checkmark & & & &\checkmark \\
2022 &Zheng et al. \cite{zheng2022autonomous} & &\checkmark & & &\checkmark & & &\checkmark \\
2022 &Yang et al. \cite{yang2022online} &\checkmark & & &\checkmark & & &\checkmark & \\
2023 &Lv et al. \cite{lv2023learning} &\checkmark & & &\checkmark & & & &\checkmark \\
\hline
\multicolumn{10}{l}{\textbf{Imitation learning}}\\
\hline
2015 &Kudoh et al. \cite{kudoh2015air} &\checkmark & & &\checkmark & & & &\checkmark \\
2017 &Yang et al. \cite{yang2016repeatable} & &\checkmark & &\checkmark & & & &\checkmark \\
2017 &Pignat and  Calinon \cite{pignat2017learning} & &\checkmark & &\checkmark & & & &\checkmark \\
2018 &Cherubini et al. \cite{cherubini2018towards} & &\checkmark & &\checkmark & & & &\checkmark \\
2020 &Seita et al. \cite{seita2020deep} & &\checkmark & &\checkmark & & & &\checkmark \\
2022 &Ha and Song \cite{ha2022flingbot} & &\checkmark & &\checkmark & & & &\checkmark \\
2022 &Avigal et al. \cite{avigal2022speedfolding} & &\checkmark & &\checkmark & & & &\checkmark \\
2022 &Lin et al. \cite{lin2022diffskill} & & &\checkmark &\checkmark & & & &\checkmark \\
2022 &Lin et al. \cite{lin2022planning} & & &\checkmark &\checkmark & & & &\checkmark \\
2023 &Ren et al. \cite{ren2023autonomous} & & &\checkmark &\checkmark & & & &\checkmark \\
\hline
\end{tabular}
\end{center}
\end{table*}

\section{Discussion of Large Languages Models and Outlooks}
LLMs recently have achieved impressive performance on various natural language processing (NLP) tasks\cite{yang2023harnessing, zhao2023survey}. Consequently, there has been a surge of interest in their potential application to the field of robotic manipulation. How can the rich internalized knowledge of the LLMs be leveraged in robotic manipulation? Recent studies utilizing LLMs have made initial progress in task definition for manipulation, planning, reward function design and uncertainty alignment. LLMs are now mainly used for rigid object manipulation. For more complex tasks involving DOM, further research is required.

\subsection{Task definition}
For accomplishing a robot manipulation task, the task first needs to be defined at the semantic level. For example, what does it mean to fold a piece of clothing? Existing approaches tend to design one configuration as the task goal. However, it is impractical to manually specify all configurations of DOs as the goal for such tasks. Instead, we need a method to acquire the semantic knowledge of concepts, such as folding or wrapping, so that we can evaluate the validity of a goal state of an object configuration. LLMs exhibit a surprising ability to generalize, providing a possible way to achieve task definitions for DOM at the semantic level.

Moreover, recent studies have made progress in learning human preferences using LLMs. A system developed by Wu \textit{et al.} \cite{wu2023tidybot} demonstrates the ability to learn human preferences by engaging with an individual using a limited number of examples. The finding indicates that robots can effectively leverage language-based planning and perception, along with the summarization capabilities of LLMs, to infer generic user preferences that can be widely applied to future interactions. The generalized user preferences could potentially serve as a priori knowledge for defining manipulation tasks. Consequently, it is worthwhile to explore task definitions that incorporate human preferences utilizing LLMs.

\subsection{Planning}
Combining abstract linguistic instructions (e.g., folding a shirt) with robot actions is an important step in accomplishing robot manipulation tasks. Previous studies have utilized lexical analysis to parse instructions \cite{chowdhery2022palm, bommasani2021opportunities, tellex2020robots}. More recent studies have utilized language models to decompose instructions into a series of literal steps \cite{huang2022language, brohan2023can, zeng2022socratic}. To achieve the physical interaction between the robot and the environment according to the processed instructions, existing methods typically depend on primitives (i.e., skills) that are either manually designed or pre-trained. For instance, a series of primitive actions such as grasping, rolling, pushing, pulling, tilting, closing, opening are pre-designed manually or trained. These primitives can be invoked by LLMs or a planner to accomplish the corresponding tasks. Liang \textit{et al.} \cite{liang2023code} showed that LLMs exhibit behavioral commonsense that can be used for underlying control. Despite some promising signs, there is still a need to manually design motion primitives. This reliance on individual skill acquisition is often considered as a significant bottleneck for the system due to the scarcity of large-scale robotic manipulation data. How can the rich internalized knowledge of the LLMs be leveraged to provide robots with a finer-grained level of action without the need for laborious data collection or manual design for each individual primitive? This research direction holds immense value.

\subsection{Reward function design}
Reward design in RL is challenging since specifying the notion of desired human behavior through a reward function may be difficult or require much expert argumentation. Consequently, an intriguing question arises: can rewards be designed using a natural language interface? Kwon \textit{et al.} \cite{kwon2023reward} explored how reward design can be simplified by prompting LLMs. In this approach, users provide textual prompts that contains several examples or descriptions of desired behaviors. The approach utilizes this agent reward functionality within the RL framework. Specifically, users specify a cue once at the beginning of training. During training, the LLM evaluates the behavior of the RL agent based on the desired behavior described by the cue and outputs a corresponding reward signal. Then, the RL agent updates its behavior using that reward. The limitation of the framework is the LLMs can only specify binary rewards. It is worth investigating how the likelihood values generated by LLMs for each word can be used as non-binary reward signals in future research. Similar works \cite{du2023guiding, ma2023eureka} have also suggested the use of pre-trained foundation models to generate reward functions for new robotic manipulation tasks. However, further research should be conducted to explore for more complex tasks involving DOM.

\subsection{Uncertainty alignment for language-instructed robots}
Large-scale language modeling has demonstrated a wide range of capabilities, from step-by-step planning to commonsense reasoning, which can be useful for robots. However, these models are still prone to confident illusory predictions. In this regard, Yu \textit{et al.} \cite{yu2023language} proposed a framework for measuring and adjusting uncertainty in language model-based planners. This framework enables robots to recognize when they lack knowledge and seek assistance when necessary. A key assumption of this framework’s task completion guarantee is that the environment (objects) is solely based on the text input to the language model. Additionally, it assumes that the actions proposed by the language model planner can be executed successfully. Future research needs to incorporate uncertainty in perceptual modules (e.g., visual language models) and underlying action strategies (e.g., language-conditioned affordance prediction).

\section{Conclusion}
In this review, we have presented a comprehensive review of recent advances, open challenges and new frontiers of DOM. We mainly focus on data-driven approaches in aspects of perception, modeling and manipulation, although analytical approaches are also concisely reviewed. 

For perception, visual perception and tactile perception methods are reviewed. The robust sensing of DOs in the presence of complex backgrounds poses an open challenge for visual perception. Tactile perception heavily depends on the use of high-resolution tactile devices, both in real-world scenarios and simulations. Additionally, we regard multimodal sensing as future research direction for perception.

For modeling, analytical modeling approaches and data-driven modeling approaches are reviewed. Analytical modeling approaches cannot accurately represent DOs due to their infinitely-dimensional states and the challenges associated with obtaining DOs' parameters in real-world settings. In contrast, data-driven approaches can generate a reasonably accurate model from data without the need for understanding the physical dynamics of DOs, given an abundance of high-quality data. Whether analytical or learned models are employed, their predictive power is limited. When the model is unreliable, it may be desirable to replan or recover. 

For manipulation, analytical manipulation methods and learning-based methods are reviewed. When the accurate dynamic model of DOs is available, analytical manipulation methods are preferred. However, obtaining the explicit dynamic model of DOs is challenging. Learning-based approaches are more versatile as they do not require the explicit dynamic model of DOs. In the future, it is crucial to explore long-horizon tasks that involve multiple tools, dexterous hands and compositional generalization.

For LLMs, recent studies have made initial progress in task definition for manipulation, planning, reward function design and uncertainty alignment. However, leveraging the rich internalized knowledge of LLMs in robotic manipulation especially for DOM still requires further investigation.

The survey provides insights into advanced research of DOM mainly focusing on data-driven approaches, and points out promising future research such as LLMs for DOM.
\newpage
\bibliographystyle{IEEEtran}
\bibliography{reference.bib}
\end{CJK}
\end{document}